\documentclass[runningheads]{llncs}

\usepackage[T1]{fontenc}
\usepackage{booktabs}
\usepackage{graphicx,verbatim}
\usepackage{graphicx}
\usepackage{booktabs}
\usepackage{tabularx}
\usepackage{multirow}

\usepackage[hidelinks,colorlinks=true,linkcolor=blue,citecolor=blue]{hyperref}
\usepackage{amsmath}
\usepackage{amssymb}

\usepackage[capitalize]{cleveref}
\crefname{section}{Sec.}{Secs.}
\Crefname{section}{Section}{Sections}
\Crefname{table}{Table}{Tables}
\crefname{table}{Tab.}{Tabs.}

\begin{document}
\title{A Multi-Center Benchmark for Abdominal Disease Diagnosis and Report Generation from Non-Contrast CT
}
\titlerunning{NCCT}
%

\author{Mariam Elbakry\inst{1} \and
Aliaa Sayed Sheha\inst{1} \and
Salma Hassan Tantawy\inst{1}\orcidID{0009-0002-1484-5883} \and
Aya Yassin\inst{1}\orcidID{0000-0001-7761-1166} \and
Concetto Spampinato\inst{3}\orcidID{0000-0001-6653-2577} \and
Karim Lekadir\inst{4}\orcidID{0000-0002-9456-1612} \and
Xiaomeng Li\inst{2}\orcidID{0000-0003-1105-8083} \and
Marawan Elbatel\inst{2}\thanks{Corresponding author: \email{mkfmelbatel@connect.ust.hk}}\orcidID{0009-0008-5021-4281}}

\authorrunning{M. Elbakry et al.}
\institute{Ain Shams University, Cairo, Egypt \\
\and
The Hong Kong University of Science and Technology, Hong Kong SAR, China \\ \and
University of Catania, Catania, Italy \\
 \and
Universitat de Barcelona, Barcelona, Spain \\
}

\maketitle              
\begin{abstract}

Multiphasic contrast-enhanced CT (CECT) is widely used for abdominal lesion characterization, yet it carries inherent risks of contrast-induced nephropathy, escalates acquisition burden, and heavily contributes to radiologist workload. To address these challenges, we introduce a novel multi-center benchmark for multi-organ abdominal disease diagnosis and automated radiology report generation, which learns to synthesize contrast-enhanced findings from single-phase non-contrast CT (NCCT). To support this, we curated a large-scale dataset of paired NCCT–CECT studies and their corresponding contrast-enhanced radiology reports from two centers, partitioned into internal sets and an external validation cohort. Under a unified evaluation protocol, we benchmarked five contemporary deep learning architectures encompassing chest-specific, abdomen-specific, and general-purpose multimodal domains. Extensive experiments demonstrate that NCCT retains diagnostic signals, achieving an average multi-organ AUC of 69.1\% on the internal cohort and 63.1\% on the external cohort, respectively. By releasing this dataset and standardized benchmark publicly, this study aims to catalyze future research into safer, resource-efficient, and globally accessible contrast-free abdominal imaging workflows. Code is available at: \url{https://github.com/xmed-lab/TriALS-Report}.

\end{abstract}
%
%
%

\section{Introduction}
Multiphasic computed tomography (CT) is the primary clinical standard for evaluating abdominal diseases, as the accurate classification of complex pathologies heavily relies on phase-specific enhancement patterns~\cite{Rydberg2000,lirads,literature_tri_ademona,litearature_tri_hcc,literature_tri_hcc,Uhm2021_literature_tri_kidney,literature_multi_phase_new_dataset_HCC_classify}. However, the use of iodinated contrast agents poses significant clinical and logistical challenges. Contrast administration is contraindicated in patients with renal impairment or hypersensitivity, and it creates substantial barriers in resource-constrained settings where economic limitations or supply chain instabilities restrict its availability. Non-contrast CT (NCCT) offers a safer, more globally accessible alternative. While the diagnostic features essential for comprehensive abdominal evaluation are often imperceptible to the human eye on NCCT alone, deep learning models have the potential to extract these latent, sub-visual morphological indicators, presenting an opportunity to bypass the need for contrast administration. Therefore, this study aims to evaluate the feasibility of generating comprehensive, contrast-enhanced radiological reports directly from single-phase pelviabdominal NCCT, utilizing ground-truth annotations derived from their contrast-enhanced counterparts.

Despite these opportunities, automated radiology report generation has predominantly been explored in thoracic imaging, with prior work heavily focused on chest X-rays and CT studies~\cite{CT2rep,btb3d,CT_CLIP_literature,literature_cxr,literatyure_cxr}. In abdominal imaging, deep learning efforts have largely been restricted to isolated diagnostic classification or lesion characterization tasks rather than comprehensive report generation~\cite{Jorg2024_literature_reports}. Furthermore, these existing approaches frequently depend on contrast-enhanced data~\cite{merlin,pillar0,total_fm_literature}. Although recent studies have explored the use of NCCT as an alternative to contrast-enhanced imaging, they have been strictly confined to single-organ or single-pathology applications, such as pancreatic cancer detection, automated urinary stone reporting, or liver lesion segmentation~\cite{literature_pancreatic_cancer,literature_renal_report_generation,elbatel2026trialstriphasicaidedliverlesion}. Crucially, the advancement of this field is severely hindered by a lack of publicly available data. Consequently, no prior work has introduced a dataset and benchmark enabling multi-organ abdominal lesion detection, characterization, and full report generation from non-contrast CT alone.

\begin{figure}[t]
\centering
\includegraphics[width=\linewidth]{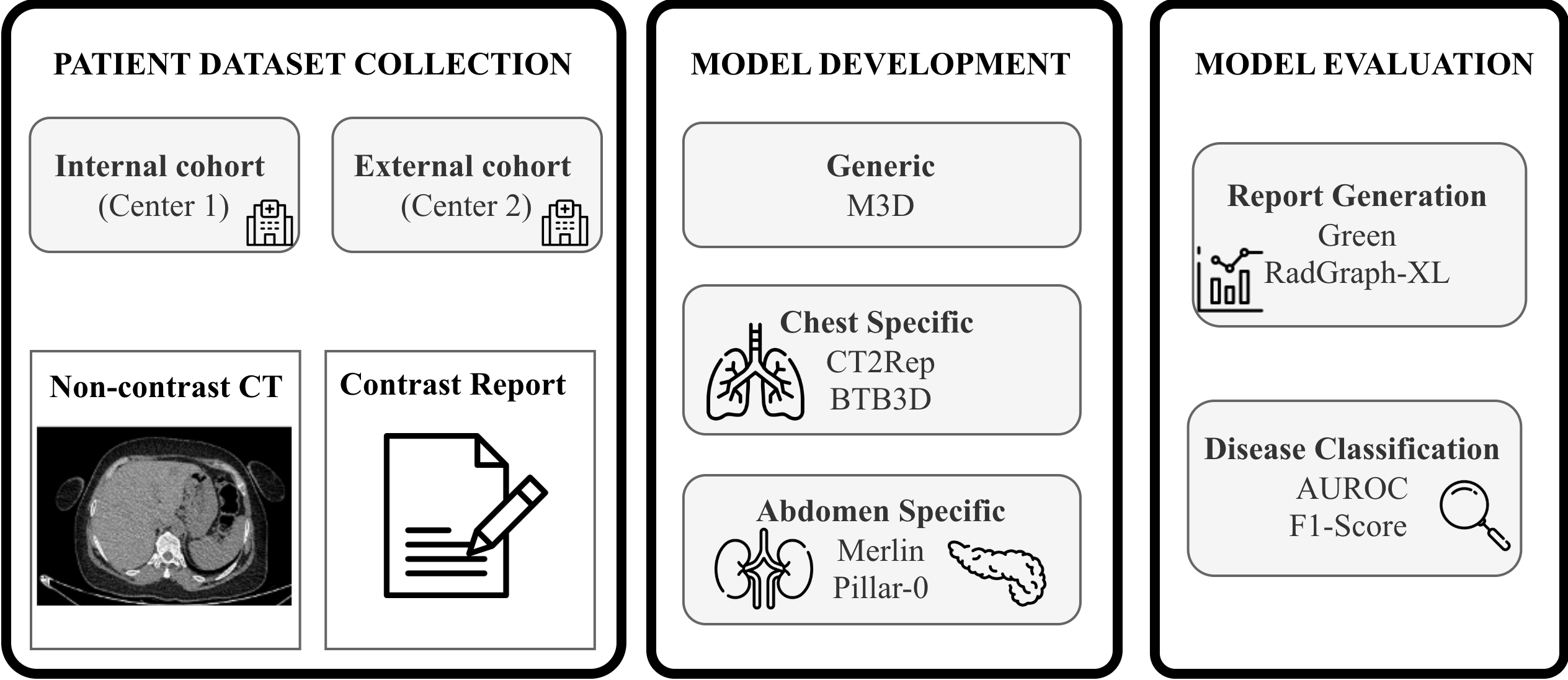}
\caption{Overview of the benchmark workflow.}
\label{fig:benchmark_main}
\vspace{-0.5cm}
\end{figure}

To bridge this critical gap, we introduce a novel dataset and comprehensive benchmark specifically designed for the automated generation of contrast-enhanced radiological reports using only non-contrast CT as the input. To the best of our knowledge, this study is among the first to evaluate the feasibility of mapping the latent features of single-phase pelviabdominal NCCT scans to complex, multi-organ disease characterizations derived from contrast-enhanced reference reports. By establishing a standardized benchmark for this task, our work provides a foundational step toward reducing radiologist workload and improving diagnostic efficiency. Ultimately, successful automated reporting from NCCT alone has the potential to significantly enhance patient safety and expand access to high-quality abdominal diagnostics in environments where multiphasic imaging is unavailable.
\section{Methodology}

\subsection{Overall Benchmark}
As illustrated in~\cref{fig:benchmark_main}, our benchmark follows a structured four-stage workflow: (1) Patient Dataset Collection, involving the curation of multi-center NCCT volumes paired with triphasic clinical ground truth; (2) Model Development, where five architectures across three paradigms are fine-tuned on the internal cohort to adapt to the Egyptian population; (3) Model Evaluation, featuring an LLM-based scoring and abdominal disease diagnosis performance. This end-to-end framework ensures that models are evaluated not only on their generative capabilities but on their ability to synthesize triphasic-level diagnostic insights from non-contrast spatial features.

\begin{table}[t]
\centering
\caption{Dataset across the participating centers. Each case consists of a non-contrast CT volume paired with a comprehensive triphasic radiology report.}
\label{tab:dataset_splits}
\resizebox{0.5\linewidth}{!}{
\begin{tabular}{@{}lccc@{}}
\toprule
\textbf{Feature} & \textbf{Internal} & \textbf{External} & \textbf{Total} \\ \midrule
Patients / Volumes & 1,085 & 169 & 1,254 \\
Train Split & 760 & -- & 760 \\
Val Split & 106 & -- & 106 \\
Test Split & 219 & 169 & 388 \\ \midrule
Matrix Size & \multicolumn{3}{c}{$512 \times 512$} \\
In-plane Spacing (mm) & \multicolumn{3}{c}{$0.875 \times 0.875$} \\
Slice Thickness (mm) & \multicolumn{3}{c}{1.07} \\ \bottomrule

\end{tabular}}
\end{table}

\subsection{Dataset}
This retrospective study was approved by the Faculty of Medicine, Ain Shams University Research Ethics Committee under protocol {FMASU MS 261/2026}. All DICOM volumes and accompanying radiology reports were de-identified before model training. Since there is a lack of publicly available triphasic CT datasets tailored for comprehensive abdominal pathology analysis, we curated a multi-center cohort comprising 1,254 patients across two tertiary institutions as depicted in~\Cref{tab:dataset_splits}. This dataset, derived from an Egyptian population, provides vital regional demographic representation. The cohort was stratified into an internal validation set (Center 1, $n=1,085$) and an external validation set (Center 2, $n=169$) to ensure model generalizability across institutional heterogeneities. Center 1 was partitioned into 760 training, 106 validation, and 219 testing cases. To maintain high spatial fidelity for capturing fine-grained pathologies, all volumes possess an axial matrix size of $512 \times 512$, characterized by a mean in-plane pixel spacing of $0.875 \times 0.875$ mm and an average slice thickness of 1.07 mm. Ground-truth diagnostic labels were extracted from the triphasic radiology reports using the RATE framework~\cite{pillar0}. To ensure clinical reliability and mitigate generative hallucinations, a subset of the extracted labels was audited by board-certified radiologists. The LLM achieved an average clinical relevance score of 90\% for disease extraction. Minor terminology variations and spelling discrepancies were manually normalized to ensure a consistent label space for model training.

\subsection{Model Development}
The evaluated models encompass three distinct archetypes: non-abdominal CT report generation frameworks, generic multimodal 3D architectures, and abdominal CT-specific models. First, we evaluate \textbf{CT2Rep}~\cite{CT2rep}, an architecture originally trained on chest CT studies that employs a 3D vision encoder coupled with a novel auto-regressive causal transformer and relational memory. Similarly, we integrate \textbf{BTB3D}~\cite{btb3d}, a causal convolutional encoder-decoder that unifies 2D and 3D training while producing frequency-aware volumetric tokens. Though optimized for the chest, its tokenization approach provides a robust baseline for high-resolution 3D medical imaging. Second, we evaluate \textbf{M3D}~\cite{m3d}, representing the generic multimodal 3D report generation paradigm. Capable of processing disparate modalities including CT, MRI, and PET, M3D utilizes a 3D spatial pooling perceiver and a large language model backbone, pretrained on multi-organ datasets. Finally, we incorporate domain-specific foundational models: \textbf{Merlin}~\cite{merlin}, a 3D vision-language model originally pre-trained on large-scale pancreatic and abdominal CT scans paired with radiology reports; and \textbf{Pillar-0}~\cite{pillar0}, a highly sample-efficient radiology foundation model trained across massive datasets of chest, abdomen, and brain volumes. Table \ref{tab:baselines} summarizes the core technical attributes of these baselines.

\begin{table}[t]
\centering
\caption{Summary of the five baseline deep-learning models, detailing their core modalities, spatial processing dimensions, and original pretraining data domains. \textit{All models have been finetuned on internal training set.}}
\label{tab:baselines}
\resizebox{\textwidth}{!}{%
\begin{tabular}{@{}llcl@{}}
\toprule
\textbf{Model} & \textbf{Modality Focus} & \textbf{2D/3D} & \textbf{Pretraining Dataset Domain} \\ \midrule
CT2Rep~\cite{CT2rep} & CT & 3D & Chest CT (Diverse populations) \\
BTB3D~\cite{btb3d} & CT & 3D & Chest CT (Large-scale corpora) \\
M3D~\cite{m3d} & Multimodal (CT, MRI, PET) & 3D & Multi-organ medical datasets \\
Merlin~\cite{merlin} & CT & 3D & Abdominal \& Pancreatic CT \\
Pillar-0~\cite{pillar0} & Multimodal (CT, MRI) & 3D & Chest, Abdomen, Head volumes \\ \bottomrule
\end{tabular}%
}
\end{table}

\subsection{Benchmark Tasks and Model Evaluation}
The benchmark formalizes two primary tasks: multi-label disease classification, identifying the presence of distinct abdominal pathologies, and automated report generation, synthesizing a coherent, free-text radiological findings. We standardize the evaluation protocol by decoupling linguistic metrics from clinical efficacy (CE) metrics to prioritize diagnostic performance.

\noindent{\textbf{Report Generation Evaluation}}. Historically, AI-generated radiology reports have been evaluated using standard natural language generation (NLG) metrics such as BLEU, ROUGE, and METEOR. However, contemporary literature demonstrates that these text-overlap metrics are systematically flawed in the medical domain, as they are heavily biased toward matching templated boilerplate text and fail to penalize catastrophic clinical errors like hallucinated findings or reversed negations. To address this, we further incorporate GREEN~\cite{green_metrics} and RadGraph-XL (Entity, Partial, and Complete)~\cite{radgraph_xl} to evaluate structured clinical entity extraction and relation correctness. Traditional NLG metrics are reported strictly as secondary, descriptive endpoints.

\noindent{\textbf{Classification Evaluation.}} For the disease classification task, models are evaluated using the Area Under the Receiver Operating Characteristic curve (AUROC) and the macro-averaged F1-score across all 53 abdominal pathologies. This multi-metric approach ensures a robust, clinical-facing evaluation of AI generalizability in a diverse, non-contrast setting despite the inherent challenges of legacy scanner protocols and retrospective data.
\begin{table*}[t]
\centering
\caption{Comparison of radiology report generation methods on internal (Center 1) and external (Center 2) datasets for non-contrast reports, evaluated via GREEN~\cite{green_metrics} and RadGraph-XL~\cite{radgraph_xl}. All methods were trained on the internal training set, except for the zero-shot baselines.}
\label{tab:report_generation_combined}
\resizebox{\textwidth}{!}{%
\begin{tabular}{l | c c c c | c | c | c | c}

\toprule
\multirow{2}{*}{ Method} &
\multirow{2}{*}{BLEU-4} &
\multirow{2}{*}{ METEOR} &
\multirow{2}{*}{ BERT} &
\multirow{2}{*}{ ROUGE-L} &
\multirow{2}{*}{ GREEN} &
\multicolumn{3}{c}{RadGraph-XL} \\
\cline{7-9}
& & & &  & & E & P & C \\
\hline


\midrule
\multicolumn{9}{c}{\textit{Internal Validation}} \\
\midrule
M3D (Zero-Shot)  & 0.056  & 6.062  & 78.451 & 8.857  & 1.695  & 4.754  & 3.857  & 1.868  \\
Merlin (Zero-Shot)& 0.537  & 13.451 & 78.513 & 13.515 & 6.254  & 19.859 & 15.154 & 12.165 \\
\hline
M3D              & 16.754 & 31.298 & 85.108 & 32.402 & 35.626 & 36.328 & 32.251 & 26.286 \\
Merlin           & 21.944 & 38.843 & 85.291 & 32.351 & 34.988 & 41.517 & 36.797 & 30.190 \\
CT2Rep           & 17.739 & 41.130 & 87.594 & 49.334 & 64.978 & 52.895 & 46.118 & 44.139 \\
BTB3D            & 27.796 & 39.183 & 85.970 & 36.137 & 36.734 & 43.451 & 38.549 & 32.079 \\
\midrule
\multicolumn{9}{c}{{\textit{External Validation}}} \\
\midrule
M3D (Zero-Shot)  & 0.072  & 5.952  & 78.116 & 8.527  & 1.279  & 4.944  & 3.970  & 1.460  \\
Merlin (Zero-Shot)& 0.594  & 17.000 & 78.543 & 13.732 & 4.967  & 19.398 & 14.358 & 11.261 \\
\hline
M3D              & 6.160  & 21.668 & 81.806 & 21.301 & 24.192 & 23.021 & 19.893 & 13.896 \\
CT2Rep           & 2.617  & 26.114 & 83.460 & 25.004 & 40.545 & 29.196 & 25.264 & 22.289 \\
Merlin           & 8.731  & 28.822 & 82.766 & 22.245 & 26.597 & 27.585 & 23.745 & 16.905 \\
BTB3D            & 6.920  & 26.813 & 82.702 & 21.983 & 29.132 & 27.495 & 23.437 & 16.349 \\
\bottomrule
\end{tabular}%
}
\end{table*}

\section{Results}
\noindent{\textbf{Contrast Free Automated Report Generation.}}The generative capabilities of the models are evaluated in~\Cref{tab:report_generation_combined}. Emphasizing clinical correctness over rudimentary text-overlap, we prioritize the GREEN and RadGraph-XL metrics. The zero-shot foundational models (M3D and Merlin) exhibit catastrophic failure when applied out-of-the-box to complex, multi-organ abdominal reporting, severely underperforming across all clinical metrics. Conversely, fine-tuning within our benchmark framework yields massive quantitative gains. We first observe this profound impact on the foundational Merlin architecture: transitioning from zero-shot inference to task-specific fine-tuning catapults its GREEN score on the internal cohort from 6.254 to 34.988, a staggering 459.4\% relative improvement. While Merlin's performance exhibits an expected degradation when exposed to the unseen external Center 2 distribution (achieving a GREEN score of 26.597), it nevertheless maintains a remarkable 435.4\% improvement over its external zero-shot baseline (4.967), underscoring the necessity of targeted adaptation for complex abdominal reporting from NCCT compared to Zero-shot.

\begin{figure}[t]
    \centering
    \includegraphics[width=\linewidth]{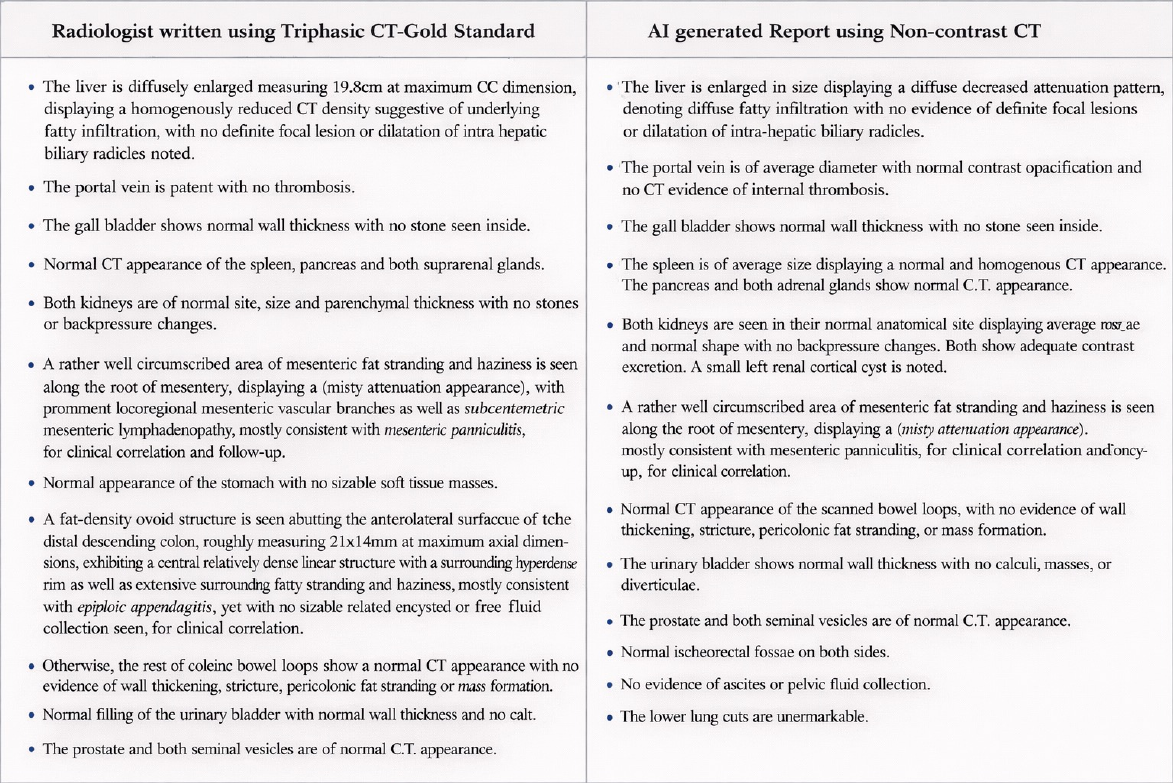}
    \caption{A comparison between AI-generated findings and expert ground truth (GT) using only non-contrast CT (NCCT) inputs.}
    \label{fig:qualiative_report}
    \vspace{-0.5cm}
\end{figure}

~\Cref{fig:qualiative_report} provides a qualitative comparison between the ground-truth report (derived from the clinical triphasic standard) and the AI-generated report utilizing strictly NCCT inputs. The qualitative alignment powerfully reinforces the quantitative gains observed in Table 3. The AI framework successfully synthesizes contrast-enhanced findings, accurately identifying diffuse fatty infiltration of the liver, normal portal vein patency without thrombosis, and subtle, localized findings such as a small left renal cortical cyst. Furthermore, the model accurately details regional inflammatory markers, correctly describing mesenteric fat stranding and haziness consistent with mesenteric panniculitis. By successfully extracting these sub-visual morphological indicators from a single-phase scan, the qualitative results confirm the framework's potential to generate highly accurate, comprehensive radiological reports while entirely bypassing the clinical risks and logistical burdens of contrast administration.

\begin{table}[htbp]
\centering
\caption{Disease diagnosis performance across different anatomical organs and imaging protocols. Multiphase indicates the late fusion of different CT phases. The reported averages cover 15 organs and a diverse, real-world distribution of 53 abdomen CT pathologies from the RATE-Evals~\cite{pillar0} taxonomy, a subset that is present in our datasets. These encompass both benign (e.g., simple cysts, hemangiomas) and malignant (e.g., hepatocellular carcinoma, adenocarcinomas) lesions. Best results per organ are highlighted in bold.}
\label{tab:lesion_detection}
\setlength{\tabcolsep}{5pt}
\begin{tabular}{llcccc}
\toprule
 & & \multicolumn{2}{c}{\textbf{Center 1 Cases}} & \multicolumn{2}{c}{\textbf{Center 2 Cases}} \\
\cmidrule(lr){3-4} \cmidrule(lr){5-6}
\textbf{Organ / Model} & \textbf{Imaging Protocol} & \textbf{AUC} & \textbf{F1 Score} & \textbf{AUC} & \textbf{F1 Score} \\
\midrule
\multicolumn{6}{l}{\textbf{Liver}} \\

\multirow{2}{*}{\hspace{3mm}Merlin\cite{merlin}} & NCCT & \textbf{0.7400} & \textbf{0.3717} & \textbf{0.7139} & \textbf{0.3575} \\
                        & Multiphase & 0.7466 & \textbf{0.3867} & \textbf{0.7788} & \textbf{0.4338} \\ \addlinespace
\multirow{2}{*}{\hspace{3mm}Pillar\cite{pillar0}} & NCCT & 0.7180 & 0.2700 & 0.6873 & 0.3301 \\
                        & Multiphase & \textbf{0.7588} & 0.2886 & 0.7147 & 0.3425 \\ \addlinespace
\multirow{2}{*}{\hspace{3mm}M3D~\cite{m3d}}    & NCCT & 0.6734 & 0.2992 & 0.6856 & 0.3469 \\
                        & Multiphase & 0.7300 & 0.3359 & 0.7171 & 0.3802 \\
\midrule
\multicolumn{6}{l}{\textbf{Spleen}} \\
\multirow{2}{*}{\hspace{3mm}Merlin\cite{merlin}} & NCCT & 0.6167 & \textbf{0.2814} & 0.6523 & \textbf{0.3406} \\
                        & Multiphase & \textbf{0.6984} & \textbf{0.2718} & \textbf{0.7464} & \textbf{0.4048} \\ \addlinespace
\multirow{2}{*}{\hspace{3mm}Pillar\cite{pillar0}} & NCCT & \textbf{0.6545} & 0.2698 & \textbf{0.7031} & 0.3021 \\
                        & Multiphase & 0.6153 & 0.2206 & 0.6207 & 0.3524 \\ \addlinespace
\multirow{2}{*}{\hspace{3mm}M3D\cite{m3d}}    & NCCT & 0.5662 & 0.1691 & 0.6109 & 0.3079 \\
                        & Multiphase & 0.6522 & 0.2392 & 0.5964 & 0.3144 \\
\midrule
\multicolumn{6}{l}{\textbf{Pancreas}} \\
\multirow{2}{*}{\hspace{3mm}Merlin\cite{merlin}} & NCCT & 0.7615 & 0.0556 & 0.6158 & \textbf{0.1766} \\
                        & Multiphase & 0.8378 & \textbf{0.1794} & 0.8630 & \textbf{0.1355} \\ \addlinespace
\multirow{2}{*}{\hspace{3mm}Pillar\cite{pillar0}} & NCCT & \textbf{0.8032} & 0.1337 & \textbf{0.7873} & 0.0814 \\
                        & Multiphase & \textbf{0.8419} & 0.1605 & \textbf{0.9289} & 0.1133 \\ \addlinespace
\multirow{2}{*}{\hspace{3mm}M3D~\cite{m3d}}    & NCCT & 0.7296 & \textbf{0.1691} & 0.5615 & 0.0000 \\
                        & Multiphase & 0.7213 & 0.1036 & 0.6990 & 0.0247 \\
\midrule
\multicolumn{6}{l}{\textbf{Kidneys}} \\
\multirow{2}{*}{\hspace{3mm}Merlin\cite{merlin}} & NCCT & 0.5861 & \textbf{0.2306} & 0.6095 & \textbf{0.2055} \\
                        & Multiphase & 0.6103 & 0.2259 & 0.6216 & 0.2136 \\ \addlinespace
\multirow{2}{*}{\hspace{3mm}Pillar\cite{pillar0}} & NCCT & \textbf{0.6373} & 0.2185 & \textbf{0.6885} & 0.1728 \\
                        & Multiphase & \textbf{0.7048} & \textbf{0.2866} & \textbf{0.7192} & \textbf{0.2876} \\ \addlinespace
\multirow{2}{*}{\hspace{3mm}M3D\cite{m3d}}    & NCCT & 0.5303 & 0.1473 & 0.4994 & 0.1798 \\
                        & Multiphase & 0.6082 & 0.2198 & 0.5645 & 0.2191 \\
\midrule
\multicolumn{6}{l}{\textbf{Average across 15 different organs and 53 diseases}} \\
\multirow{2}{*}{\hspace{3mm}Merlin\cite{merlin}} & NCCT & \textbf{0.6910} & \textbf{0.2513} & \textbf{0.6312} & \textbf{0.2477} \\
                        & Multiphase & \textbf{0.7654} & \textbf{0.2851} & \textbf{0.6861} & \textbf{0.2560} \\ \addlinespace
\multirow{2}{*}{\hspace{3mm}Pillar\cite{pillar0}} & NCCT & 0.6901 & 0.2153 & 0.6271 & 0.2022 \\
                        & Multiphase & 0.7382 & 0.2348 & 0.6671 & 0.2385 \\ \addlinespace
\multirow{2}{*}{\hspace{3mm}M3D~\cite{m3d}}    & NCCT & 0.6616 & 0.2147 & 0.5925 & 0.1988 \\
                        & Multiphase & 0.6965 & 0.2302 & 0.6243 & 0.2356 \\
\bottomrule
\end{tabular}
\end{table}

\noindent{\textbf{Disease Diagnosis Performance}}
The diagnostic classification capabilities of the evaluated architectures across 15 organs and 53 distinct abdominal pathologies are detailed in~\Cref{tab:lesion_detection}. As anticipated, models utilizing multiphasic contrast-enhanced inputs establish the upper performance bound, reflecting the rich temporal enhancement dynamics inherent to the clinical gold standard. Merlin achieves the highest overall diagnostic performance on multiphasic data, securing an average AUC of 0.7654 and 0.6861 on the internal and external cohorts, respectively. Crucially, when restricted to single-phase NCCT inputs, the models demonstrate remarkable diagnostic retention and highly competitive zero-contrast classification capabilities. Across the 15-organ average, Merlin’s NCCT performance (AUC 0.6910) retains over 90\% of its multiphasic diagnostic power (AUC 0.7654). We observe similarly robust retention at the organ-specific level; for instance, Pillar achieves an impressive AUC of 0.8032 for pancreatic lesion detection on NCCT (Center 1), effectively bridging the gap to complex triphasic detection rates. Furthermore, Merlin achieves a robust AUC of 0.7400 for liver pathologies from NCCT inputs, compared to 0.7466 using multiphasic data. This high retention rate across a diverse taxonomy provides compelling empirical evidence that deep learning frameworks can successfully map latent non-contrast features to complex, multi-organ disease characterizations, validating the clinical feasibility of NCCT-driven diagnostics.

\section{Discussion} 

This study introduces the first multi-centric dataset and benchmark dedicated to contrast-free abdominal disease diagnosis and automated report generation. By successfully mapping latent non-contrast spatial features to complex triphasic-equivalent findings, our benchmark proves that AI-driven NCCT retains ~90\% of AI-driven CECT diagnostic power. Furthermore, the massive quantitative leaps achieved over zero-shot foundational models decisively validate the necessity of our targeted, multi-centric adaptation pipeline for complex abdominal reporting.

However, these findings exist within specific foundational boundaries. First, while performance naturally shifts on the external Center 2 cohort due to scanner variations, this deliberately establishes a realistic, rigorous baseline that invites targeted domain adaptation research. Second, although our scalable framework relies on global text-alignment rather than pixel-level lesion grounding, this intentional design enabled massive multi-centric curation, with expert radiologist audits successfully mitigating label noise to ensure high clinical validity.

Finally, NCCT imposes a fundamental information ceiling. Enhancement-kinetic findings such as arterial hyperenhancement, portal-venous washout, delayed retention, and capsule appearance depend on intravenous contrast dynamics and cannot be physically inferred from a single-phase non-contrast scan. Within this constraint, our benchmark shows that latent morphological signatures recover the majority of AI-driven multiphasic diagnostic performance, with the residual gap concentrated on low-contrast targets such as the pancreas, where enhancement timing is most informative. We therefore position NCCT-driven reporting as a complementary triage and screening modality, particularly in settings where contrast administration is contraindicated or unavailable, rather than a replacement for multiphasic protocols. Downstream contrast-enhanced imaging remains warranted for definitive lesion characterization and for findings whose diagnosis depends on enhancement kinetics. We anticipate that the public release of this dataset and benchmark will catalyze future research into contrast-free imaging, ultimately accelerating the routine clinical deployment of safer, more globally accessible, and resource-efficient diagnostic workflows.

\begin{credits}
\subsubsection{\ackname} The work has been partially supported by the European Union – Next Generation EU, Mission 4 Component 2 Line 1.3, through the PNRR MUR project PE0000013 – FAIR “Future Artificial Intelligence Research” (CUP E63C220\allowbreak01940006). Marawan Elbatel is supported by the Hong Kong PhD Fellowship Scheme (HKPFS) from the Hong Kong Research Grants Council (RGC), and partially supported by a research grant from the Bridge Gap Fund (Project No. BGF.020.2025) at HKUST.

\subsubsection{\discintname}
The authors have no competing interests to declare that are relevant to the content of this article.
\end{credits}

%
%
\bibliographystyle{splncs04}
\bibliography{mybib}
\end{document}